\documentclass[preprint,12pt]{elsarticle}
\usepackage{amsmath,amssymb,amsfonts}
\usepackage[USenglish]{babel}
\usepackage[utf8]{inputenc}
\usepackage[T1]{fontenc}

\usepackage{array, booktabs}
\usepackage{multirow}
\usepackage{makecell}
\usepackage{adjustbox}

\usepackage{caption}
\usepackage{subcaption}

\usepackage{textcomp}
\usepackage[table,xcdraw]{xcolor}
\usepackage[hyphens]{url}

\usepackage{graphicx}
\graphicspath{{./}}

\usepackage[skins]{tcolorbox}
\usepackage{fontawesome5}

\usepackage[
colorlinks,
linkcolor={red!50!black},
citecolor={blue!50!black},
urlcolor={blue!80!black}
]{hyperref}

\hypersetup{
urlcolor={black}
}

\usepackage[toc,nopostdot,nonumberlist,acronym]{glossaries}
\glsdisablehyper

\makeglossaries

\newacronym[shortplural={DNNs}, longplural={Deep Neural Networks}]{dnn}{DNN}{Deep Neural Network}

\newacronym{ai}{AI}{Artificial Intelligence}

\newacronym{dl}{DL}{Deep Learning}

\newacronym{ml}{ML}{Machine Learning}

\newacronym[shortplural={GPUs}, longplural={Graphical Processing Units}]{gpu}{GPU}{Graphical Processing Unit}

\newacronym[shortplural={CPUs}, longplural={Central Processing Units}]{cpu}{CPU}{Central Processing Unit}

\newacronym{tdp}{TDP}{Thermal Design Power}

\newacronym[shortplural={VAEs}, longplural={Variational Autoencoders}]{vae}{VAE}{Variational Autoencoder}

\newacronym{flops}{FLOPs}{Floating-point Operations}
\newacronym[shortplural={MACs}, longplural={Multiply-accumulate operations}]{mac}{MAC}{Multiply-accumulate operation}

\newacronym{mlco2}{MLCO2}{ML CO$_2$ Impact Calculator}

\newacronym{ga}{GA}{Green Algorithms}

\newacronym{pue}{PUE}{Power Usage Effectiveness}

\newacronym{psf}{PSF}{Pragmatic Scaling Factor}

\newacronym[shortplural=RQs, longplural=Research Questions]{rq}{RQ}{Research Question}

\newacronym{hpc}{HPC}{High-Performance Computing}

\newacronym{ram}{RAM}{Random Access Memory}

\newacronym{rmse}{RMSE}{Root Mean Squared Error}

\newacronym{mae}{MAE}{Mean Absolute Error}

\newacronym[shortplural={LMs}, longplural={Linear Models}]{lm}{LM}{Linear Model}

\newacronym[shortplural={GLMs}, longplural={Generalized Linear Models}]{glm}{GLM}{Generalized Linear Model}

\newacronym{gamlss}{GAMLSS}{Generalized Additive Models for Location, Scale, and Shape}

\journal{Computer Standards \& Interfaces}

\begin{document}
\newcommand{\co}{CO$_2$}

\newcommand{\coeq}{\co{}eq}

\newcommand{\emissions}{\co{} emissions}

\newcommand{\mlco}{ML \co{} Impact Calculator}

\begin{frontmatter}

\title{Estimating Deep Learning energy consumption based on model architecture and training environment
 \tnoteref {t1}
}
\tnotetext[t1]{This work is part of the GAISSA project (TED2021-130923B-I00), which is funded by MCIN/AEI/10.13039/501100011033 and by the European Union ``NextGenerationEU''/PRTR, and the GAISSA-Optimizer research project under the AGAUR 2025 program (PROD 00236). This work is also partially supported by the Joan Or{\'o} pre-doctoral support program (BDNS 657443), co-funded by the European Union.}

\author[1]{Santiago {del Rey}}
\ead{santiago.del.rey@upc.edu}

\author[2]{Lu{\'i}s Cruz}
\ead{L.Cruz@tudelft.nl}

\author[1]{Xavier Franch}
\ead{xavier.franch@upc.edu}

\author[1]{Silverio Mart{\'i}nez-Fern{\'a}ndez\corref{cor1}}
\ead{silverio.martinez@upc.edu}

\cortext[cor1]{Corresponding author}

\affiliation[1]{
    organization={Universitat Polit{\`e}cnica de Catalunya},
    city={Barcelona},
    country={Spain}
}

\affiliation[2]{
    organization={Delft University of Technology},
    city={Delft},
    country={Netherlands}
}

\begin{abstract}
To raise awareness of the environmental impact of deep learning (DL), many studies estimate the energy use of DL systems. However, energy estimates during DL training often rely on unverified assumptions. This work addresses that gap by investigating how model architecture and training environment affect energy consumption. We train a variety of computer vision models and collect energy consumption and accuracy metrics to analyze their trade-offs across configurations. Our results show that selecting the right \textit{model--training environment} combination can reduce training energy consumption by up to 80.68\% with less than 2\% loss in $F_1$ score. We find a significant interaction effect between model and training environment: energy efficiency improves when GPU computational power scales with model complexity. Moreover, we demonstrate that common estimation practices, such as using FLOPs or GPU TDP, fail to capture these dynamics and can lead to substantial errors. To address these shortcomings, we propose the Stable Training Epoch Projection (STEP) and the Pre-training Regression-based Estimation (PRE) methods. Across evaluations, our methods outperform existing tools by a factor of two or more in estimation accuracy.
\end{abstract}

\begin{graphicalabstract}
\includegraphics[width=\textwidth]{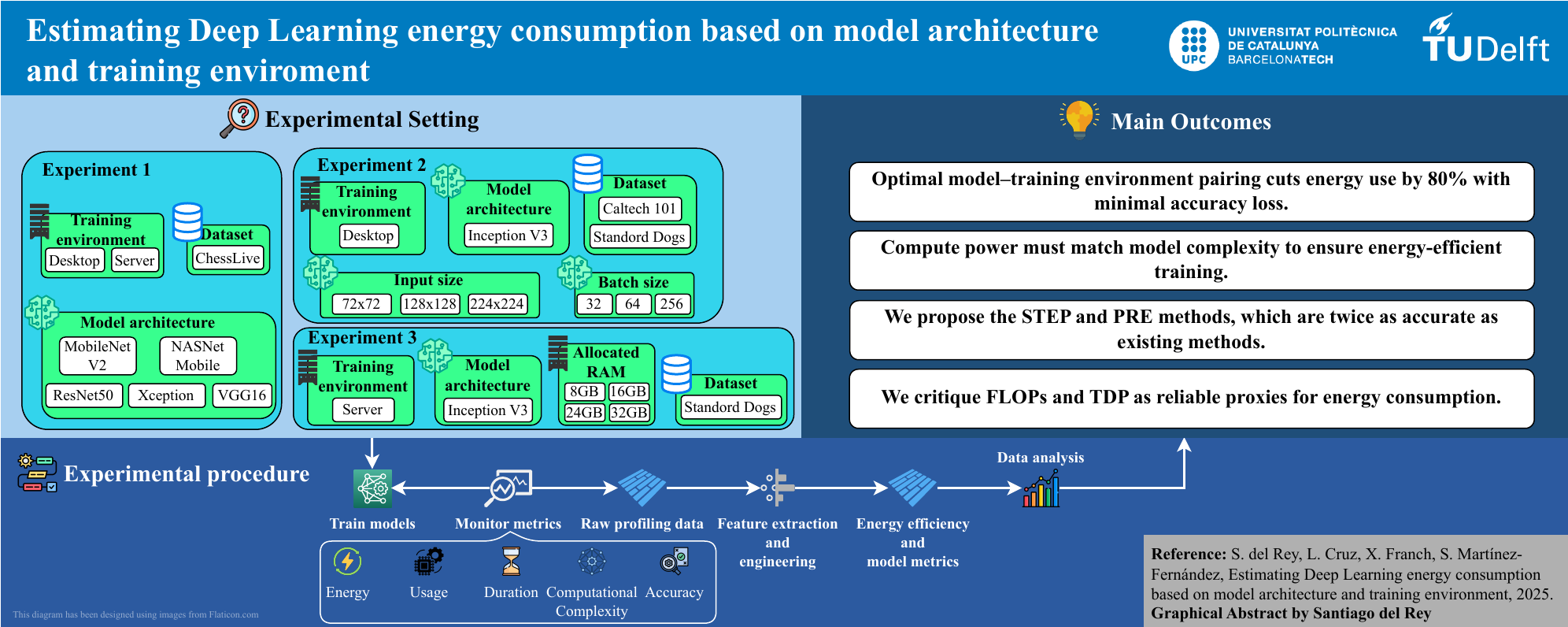}
\end{graphicalabstract}

\begin{highlights}
\item Optimal model--training environment pairing cuts energy use by 80\% with minimal accuracy loss.
\item To ensure energy-efficient training, compute power must match model complexity.
\item We propose the STEP and PRE methods, which are twice as accurate as existing methods.
\item We critique FLOPs as an energy proxy and the reliability of TDP-based estimates.
\end{highlights}

\begin{keyword}
Green IN AI \sep energy consumption estimation \sep neural networks \sep empirical software engineering \sep sustainable computing
\end{keyword}
\end{frontmatter}

\section{Introduction}
We are in a moment marked by an unprecedented surge in the utilization of \gls{ai} technologies~\cite{NBERw32966, simaremareTheStateOfGenerativeAIAdoption2024}. This surge is driven by the birth of groundbreaking technologies such as large language models and generative \gls{ai}, which have revolutionized the landscape of \gls{ai} applications~\cite{aleneziAiDrivenInnovationsInSoftwareEngineering2025}. However, this progress comes at a cost: the growing energy footprint of \gls{dl} model training~\cite{Amodei2018, patterson_carbon_2022}. For instance, it is estimated that training Gemini Ultra emitted 37,620 tonnes of \coeq~\cite{yuRevisitEnvironmentalImpact2024}, which is equivalent to the yearly emissions of 7,840 US houses' electricity use.\footnote{The equivalence has been obtained from the \href{https://www.epa.gov/energy/greenhouse-gas-equivalencies-calculator}{U.S. Environmental Protection Agency}.} As these systems become increasingly integrated into mainstream software applications, building energy-efficient training pipelines has become a pressing priority, not just for sustainability but also for scalability and cost-efficiency. Therefore, the research community is taking measures to provide novel results while considering their energy efficiency, a topic coined by \citet{Schwartz2020} as \emph{Green \gls{ai}}. Furthermore, Gartner identifies ``energy-efficient computing'' as one of the top strategic tech trends that will help CIOs and IT leaders shape the future with responsible innovation in 2025\footnote{\url{https://www.gartner.com/en/articles/top-technology-trends-2025}}.

Given the growing body of knowledge on \emph{Green \gls{ai}}, \citet{gutierrez_green_2024} propose to divide this research topic into two subareas depending on the focus of study. First, we find \textit{Green BY \gls{ai}} research, which focuses on using \gls{ai} as a medium to improve environmental sustainability in any context. Second, we find \textit{Green IN \gls{ai}} research, which deals with improving the energy consumption of \gls{ai} on itself. For instance, research aiming to find a more energy-efficient \gls{dl} model architecture is Green IN, while using a \gls{dl} model to save energy in a smart building is Green BY. This work falls under the subarea of \textit{Green IN \gls{ai}}.

The research community has made progress in proposing lightweight architectures~\cite{chenReviewLightweightDeep2024} and optimization strategies~\cite{congReviewConvolutionalNeural2023}, providing data scientists with methods and techniques to reduce training energy consumption. The role of data scientists in selecting or designing efficient models, often with pre-training and fine-tuning, is relatively well established. However, less attention has been paid to the role of software engineers, who are typically responsible for architecting the broader systems in which these models are trained.

In particular, it remains unclear what design decisions software engineers can take to reduce energy consumption. For example, should training be executed on consumer-grade desktops, cloud servers, or \gls{hpc} infrastructure? How do these environments interact with different model architectures in terms of energy consumption, and how should software engineers reason about these trade-offs in practice? From a software perspective, these are fundamental questions that remain largely unanswered in current literature and tooling.

Several standards are currently under development to define the terminology and methodology for reporting and measuring energy consumption. Recently, the International Organization for Standardization (ISO) published the ISO/IEC DTR 20226 standard~\cite{ISOIECTR}, which provides an overview of the environmental sustainability aspects (e.g., resource and asset utilization, location) of AI systems during their life cycle, and related potential metrics. The Green Software Foundation (GSF) also published the Software Carbon Intensity (SCI) Specification~\cite{SoftwareCarbonIntensity}, which defines a methodology for calculating the rate of carbon emissions for a software system. There are also national initiatives for standardizing environmental sustainability aspects, such as the Spanish Association for Standardization (UNE) working group CTN-UNE 71/SC 42/GT 1~\cite{UNEIA}, or the German Institute for Standardization (DIN) working group on resource-efficient software~\cite{DINResourceefficientSoftware}. However, as they are relatively new or still under development, their adoption is scarce.

Although several works have estimated the energy consumption, or carbon emissions, of training \gls{dl} models~\cite{patterson2021carbon, luccioniEstimatingCarbonFootprint2023}, these estimations are taken from tools that rely on assumptions that do not account for the complex interplay between hardware, software, and training configuration. As a result, practitioners are often left without reliable guidance for making energy-aware choices during \gls{dl} training pipeline design.

This paper addresses these challenges through a detailed empirical study of the interaction between model architecture and training environment. We analyze how the selection of \gls{dnn} architecture and training environment can improve the energy efficiency of \gls{dl} training pipelines without great losses in accuracy. We collect a set of energy-related metrics during the models' training. Then, we analyze their relationship with the models' architectures and the training environments selected. Furthermore, we utilize all the collected data to analyze the power behavior during the training process and utilize the knowledge obtained to propose four new energy estimation methods. To provide more precise results, we focus on \gls{dl} for computer vision.

Our results can help practitioners optimize their training pipelines, use resources more efficiently, and obtain better energy estimations. Furthermore, cloud providers could improve energy efficiency by dynamically selecting the \glspl{gpu} of a training instance depending on the model to be trained.

The rest of the document is structured as follows. Section~\ref{sec:related-work} reviews related work. Section~\ref{sec:methodology} outlines the study design and methodology. Section~\ref{sec:results} presents the results of the study, and Section~\ref{sec:discussion} discusses related points. Section~\ref{sec:threats} presents possible threats to the validity of the study and how we mitigate them. Finally, Section~\ref{sec:conclusion} concludes with the main contributions of our work and presents future work.

\section{Related Work}\label{sec:related-work}
This section starts by presenting relevant studies analyzing different design decisions that impact the energy consumption of \gls{dl} training. Then, it presents relevant methods and tools currently used to measure and estimate the energy consumption and carbon footprint in \gls{dl} training.

\subsection{Energy consumption of DL training}
The energy efficiency of \gls{dl} systems is becoming a rising research topic, encompassing the life cycle of \gls{dl} models~\cite{verdecchia2023systematic}. Most existing studies focus on comparing the energy consumption of various \gls{dl} models and training techniques. However, few works investigate how design decisions, such as the selection of model architecture or training environment, affect energy consumption.

\citet{luckow_deep_2016} examine the trade-offs between training \gls{dl} models locally versus in the cloud, reporting that cloud-based training takes approximately 1.5$\times$ longer. They also find that while multi-\gls{gpu} training reduces training time, efficiency rapidly drops. In contrast, our study focuses on single-\gls{gpu} training and factors in the energy consumption in the trade-off analysis. Additionally, we explore the interplay between \gls{gpu} efficiency and model architecture.

\citet{Li2016} conduct a comprehensive analysis of the power behavior and energy efficiency of popular \gls{dl} models and training frameworks, covering multiple hardware setups and software libraries on both \glspl{cpu} and \glspl{gpu}. Their findings indicate that convolutional layers are the primary contributors to energy usage. While their study analyzes energy at the layer level, we instead evaluate energy consumption at the network level---particularly relevant for the use of pre-trained models where access to individual layers may be limited. We further explore the trade-off between energy and accuracy, and consider three different training environments to reflect the practices of diverse user profiles.

In a similar direction, \citet{caspart_precise_2022} compare energy consumption across heterogeneous \gls{ai} workloads on varying compute nodes. They report that image classification tasks benefit from lower power draw and shorter runtimes on a single \glspl{gpu} compared to multi-core \glspl{cpu}. Moreover, they observe that \gls{gpu} idle time results in a non-negligible portion of energy consumption, recommending \gls{gpu} usage for \gls{dl} workflows when available. We extend their findings by considering five \gls{dl} models, rather than two, to improve generalizability, and examine the influence of model selection on accuracy. Additionally, we assess the impact of changing the training environment, a dimension not previously explored to our knowledge.

In the context of computer vision, \citet{jha_ramifications_2019} evaluate the implications of using compact \glspl{dnn} on memory footprint, energy efficiency (in \glspl{mac}/Joule), and throughput. They find that reducing the \glspl{mac} does not necessarily translate into lower energy consumption and higher throughput. A strong linear correlation is observed between \glspl{mac}/activation ratio and energy efficiency. We complement this work by introducing an energy-versus-accuracy trade-off analysis and examining how model architecture and training environment interact, as they are relevant design aspects for practitioners selecting or developing models.

\citet{yarallyUnvoceringEnergy2023} analyze the energy impact of convolutional, linear, and ReLU layers during CNN training. Their study confirms the high computational cost of convolutional layers and recommends minimizing their use where possible, as additional model complexity yields only marginal accuracy gains. We contribute to their work by studying the problem from a model-wise perspective rather than a layer-wise perspective. Our goal is to support practitioners who rely on pre-trained models, offering insights without requiring full model design.

Finally, \citet{regueroEnergyefficientNeuralNetwork2025} investigate the impact of design decisions (layer freezing, quantization, and early stopping) on VGG-16, and propose a predictive model that forecasts accuracy and energy consumption to identify the optimal epoch to apply these decisions. We take a related but broader approach by examining model architecture and training environment as key design factors. As emphasized by \citet{cruzGreeningAIenabledSystems2025}, sustainable \gls{dl} requires not only efficient models but also holistic, lifecycle-aware design choices. Our study directly contributes to this agenda by empirically evaluating the interplay of model architecture, training environment, and energy–accuracy trade-offs, providing evidence to inform such system-level decision making.

\subsection{Measuring energy consumption and carbon footprint in DL training}
As awareness of \gls{dl}'s environmental impact grows, research focused on estimating or measuring \gls{dl} energy consumption and carbon emissions is also increasing. Platforms like Hugging Face\footnote{\url{https://huggingface.co/docs/hub/model-cards-co2}} and the Green Software Foundation\footnote{\url{https://if.greensoftware.foundation/}} have begun to promote carbon reporting practices, reinforcing the need for reliable energy consumption measurement tools. These tools fall into two main categories: hardware-based and software-based~\cite{bouzaHowEstimateCarbon2023, guldnerDevelopmentEvaluationReference2024}. In this work, we focus on software-based tools since they tend to be more accessible.

Among software-based tools, the most popular methods are: (i) real-time energy estimators that monitor hardware usage during training and aggregate this data to compute energy and emissions, and (ii) post-training estimators, which require users to input training details and rely on assumptions about power behavior to estimate energy usage and emissions.

Regarding real-time energy estimators, CodeCarbon\footnote{\url{https://github.com/mlco2/codecarbon}} is a widely adopted tool in this space, offering an easy-to-use Python interface to monitor energy and carbon emissions. Alternatives such as PowerJoular~\cite{noureddine-ie-2022} or EnergyBridge~\cite{sallou2023energibridgeempoweringsoftwaresustainability} provide similar functionality without requiring Python. While these tools aim to continuously track energy consumption, we also explore their potential for projecting DL training energy consumption---thus saving time and computational resources.

Well-known post-training estimators include \gls{mlco2} calculator~\cite{lacoste2019quantifying}, and \gls{ga}~\cite{lannelongueGreenAlgorithmsQuantifying2021} tool. Although their primary goal is to estimate carbon emissions, they first estimate energy consumption. The \gls{mlco2} calculator assumes the \gls{gpu}'s \gls{tdp} as average power consumption, then multiplies it by execution time. \gls{ga} extends this approach by: (i) scaling \gls{tdp} based on the number of processing units and usage; (ii) including memory power draw; (iii) factoring in \gls{pue}, if known; and (iv) using the \gls{psf} to account for multiple identical runs. In this work, we propose a data-driven alternative to \gls{tdp}-based estimations, aiming for more accurate predictions.

Many studies still rely on the same foundational assumptions as \gls{mlco2} or \gls{ga}~\cite{luccioniEstimatingCarbonFootprint2023, luccioni2023counting}, without assessing their reliability. To fill this gap, we investigate power usage patterns during training to inform more robust energy estimators. This represents a novel contribution toward building sustainable and efficient computer vision systems.

\section{Study design and execution}\label{sec:methodology}
\begin{figure}[ht]
    \centering
    \includegraphics[width=\textwidth]{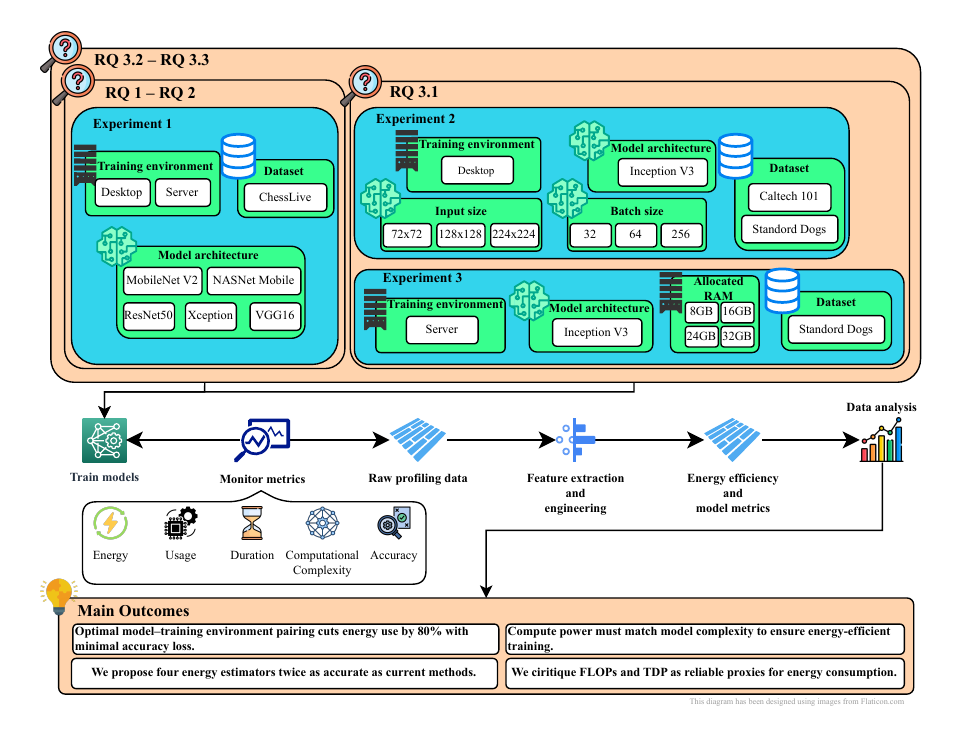}
    \caption{Structure of the study and main outcomes.}\label{fig:study-design}
\end{figure}

This section presents the context of our study, as well as our research goal and research questions. It then describes the variables of the study as well as the steps followed to perform the experiments and the posterior data analysis. Figure~\ref{fig:study-design} provides an overview of the main aspects of the study.

\subsection{Context: image classification}

We conduct our study in the context of a computer vision task: image classification. We perform our experiments on three datasets. We have selected two popular image classification datasets, namely Caltech-101 and Stanford Dogs, which provide a common benchmark for comparison with other studies. The third dataset was created by the first author for training a \gls{dl} model to detect empty squares on a chessboard. This last dataset has been used to analyze the effects of the two variables of interest in a considerably larger dataset.

\subsection{Research Goal and Questions}\label{sec:RQs}
Following the Goal-Question-Metric guidelines~\cite{Basili1994TheGQ}, we define our goal as:

{\em\noindent\textbf{Analyze} \gls{dl} model architectures and training environments\newline
\textbf{For the purpose of} estimating energy consumption\newline
\textbf{With respect to} the combined energy consumption of the \gls{gpu} and \gls{ram}, \gls{gpu} usage, model accuracy, and their trade-off\newline
\textbf{From the point of view of} data scientists and software engineers\newline
\textbf{In the context of} image classification model training.}

From this goal, we derive the following \glspl{rq}:

\begin{tcolorbox}[colback=gray!5!white, colframe=gray!60!black, title=\faQuestionCircle~Research Question 1]
What is the impact of model architecture and training environment on image classification training concerning energy consumption and \gls{gpu} usage?
\end{tcolorbox}

As we have shown, previous work has already explored how the model architecture impacts the energy consumption of training \gls{dl} models. However, there is still little evidence on the effects of the training environment and how it interacts with the model architecture. With \textit{\gls{rq}$_1$}, we aim to investigate this interaction and its impact on the combined training energy consumption of the \gls{gpu} and \gls{ram}, as well as \gls{gpu} usage. Our goal is to see if the selection of a model architecture can potentially affect the selection of the training environment when aiming to reduce the energy consumption of training.

Moreover, seeing how previous work uses the \gls{gpu} usage as a scaling factor when estimating the energy consumption~\cite{lannelongueGreenAlgorithmsQuantifying2021}, in \textit{\gls{rq}$_1$} we also study whether this assumption holds when changing the training environment. Additionally, we analyze if there is a relationship between \gls{gpu} usage and training energy consumption at all. To achieve these goals, we divide \textit{\gls{rq}$_1$} into two subquestion:

\textbf{\gls{rq}$_{1.1}$:} How do model architecture and training environment impact training energy consumption and \gls{gpu} usage?

\textbf{\gls{rq}$_{1.2}$:} Is there a relationship between \gls{gpu} usage and energy consumption when training \gls{dl} models?

To answer \gls{rq}$_1$, we measure the combined energy consumption of the \gls{gpu} and \gls{ram}. Specifically, we measure their energy consumption, in Joules, while training multiple model architectures in two different training environments (i.e., Server and Desktop, see below), and sum them to obtain their combined energy consumption for training, which we abbreviate in the remainder of this document as \textit{energy consumption}. The \gls{gpu} usage is measured as the percentage of time at least one core of the \gls{gpu} is working.

We refer to \textit{Server} environments as those present in \gls{hpc} centers or similar, which contain clusters of highly specialized hardware such as very powerful \glspl{cpu} and \glspl{gpu}, and potent refrigeration systems. For \textit{Desktop} environments, we refer to the hardware components found in a consumer-class laptop or desktop computer (e.g., medium to low-end \glspl{cpu} and \glspl{gpu}).

\begin{tcolorbox}[colback=gray!5!white, colframe=gray!60!black, title=\faQuestionCircle~Research Question 2]
Do potential gains in accuracy achieved by energy-greedy \gls{dl} model architectures justify their increased energy consumption in image classification tasks?
\end{tcolorbox}

Previous work has raised concerns regarding the current trend of increasing the computational demands of \gls{dl} models to increase their accuracy~\cite {Schwartz2020, Zhang2022, sevilla_compute_2022}. In \gls{rq}$_2$, we analyze the trade-off between energy consumption and model accuracy across five model architectures with varying computational complexity. We aim to examine whether model architectures with higher energy consumption perform better in accuracy and whether the increase in accuracy is significant, so that it can justify their higher energy demands.

To answer this \gls{rq}, we measure the energy consumption, as defined in \gls{rq}$_1$, and the model accuracy in terms of $F_1$ score.

\begin{tcolorbox}[colback=gray!5!white, colframe=gray!60!black, title=\faQuestionCircle~Research Question 3]
Can we accurately estimate energy consumption before training \gls{dl} models for image classification?
\end{tcolorbox}

With \textit{\gls{rq}$_3$} we want to develop a method that can reliably estimate energy consumption. For this, we first investigate how different parameters (i.e., batch size, input size, and allocated \gls{ram}) impact different variables like energy consumption and \gls{ram} usage. Then, we look for patterns in the power consumption of training runs to see if it is something we could use to estimate energy. Finally, we evaluate our proposed method by comparing its performance to two popular estimation methods (i.e., \gls{mlco2} and \gls{ga}). The final goal of this \gls{rq} is to provide data scientists with a way to observe the energetic implications of building their \gls{dl} models without actually training them. This way, they can compare different model architectures and training environments, in terms of energy consumption, much faster and without consuming as much energy as they would need otherwise. To achieve this goal, we divide \textit{\gls{rq}$_3$} into three subquestions:

\textbf{\gls{rq}$_{3.1}$:} Which parameters can help estimate the energy consumption?

\textbf{\gls{rq}$_{3.2}$:} Are there recurring patterns between epochs within the same training run? If so, can we use them to predict energy consumption?

\textbf{\gls{rq}$_{3.3}$:} How reliable is the proposed energy estimation method?

\subsection{Variables}\label{sec:variables}

This section presents the collection of independent and dependent variables we use for the study (see Table \ref{tab:variables}).

\textbf{Independent Variables.} We use five independent variables: model architecture, training environment, allocated \gls{ram}, batch size, and input size. Model architecture is used to describe different \gls{dl} models widely adopted by the community, which purposely vary in terms of size, type of layers, and how these layers are connected. The training environment represents possible environments that data scientists could choose when deciding where to train their models. We define one Server environment and two Desktop environments. The  ``Normal User'' Desktop environment (\textit{Desktop N}) utilizes a low-end \gls{gpu} that could be used by the average person with a computer. The ``\gls{ml} Engineer'' Desktop environment (\textit{Desktop \gls{ml}}) uses a medium-nearly-high-end \gls{gpu} more oriented to intensive computing tasks (e.g., gaming, \gls{dl} experiments). The allocated \gls{ram} variable represents the situation when data scientists must decide the amount of \gls{ram} to allocate when sending a job to train a \gls{dl} model in a compute cluster. Finally, batch size and input size are common parameters of a \gls{dl} model that must be adjusted when defining a \gls{dl} training pipeline. Indeed, batch size has been demonstrated to be useful in reducing energy consumption~\cite{edelmanAnalysisEnergyRequirement2023, alizadehGreenAIPreliminary2024}. Instead, input size increases the number of parameters of the \gls{dnn}, potentially increasing its computational demands.

\textbf{Dependent Variables.} To analyze the contribution of the independent variables to energy efficiency and accuracy, we consider six dependent variables. To evaluate energy efficiency, we measure energy consumption, \gls{gpu} usage, the \gls{ram} used during the training, the training duration, and the model computational complexity as the number of \gls{flops} required for a single forward pass of the model. For model accuracy, we measure the $F_1$ score. When looking at the results, it is important to consider that the \gls{gpu} usage reported by nvidia-smi\footnote{\url{https://developer.nvidia.com/system-management-interface}} is the percentage of time the \gls{gpu} is actively performing computations, not the percentage of cores that are being used.

\textbf{Other Variables.} To get measurements that align as much as possible with a real use case, we add the dataset as an additional variable of our study. We use the dataset to represent common tasks that image classification models are trained on. We use two popular datasets (i.e., Caltech101 and Stanford Dogs), and a third dataset we built for other research purposes. We give details of their composition in the following section.

\begin{table}[!t]
\centering
\caption{The variables of the study.}
\label{tab:variables}
\rowcolors{2}{gray!20}{white}
\small
\begin{tabular}{llp{7.5cm}}
\toprule
\textbf{Name}           & \textbf{Scale}     & \textbf{Operationalization}     \\ 
\midrule
\rowcolor{gray!60}
\multicolumn{3}{l}{Independent variables:} \\
Model architecture      & nominal   & Base model: MobileNet V2, NASNet Mobile, ResNet-50, Xception, VGG-16, Inception V3\\
Training environment    & nominal   & Three environments: Desktop "Normal User", Desktop "\gls{ml} Engineer", and Server\\
Allocated \gls{ram}     & ordinal   & Amount of \gls{ram} allocated for the training process (8GB, 16GB, 24GB, 32GB)\\
Batch size              & ordinal   & Number of images processed in a training step (32, 64, 256)\\
Input size              & ordinal   & Width, and height in pixels of the input images (72x72, 128x128, 224x224)\\
\rowcolor{gray!60}
\multicolumn{3}{l}{Dependent variables:} \\
Energy consumption      & ratio & Profiled with nvidia-smi and psutil\\
\gls{gpu} usage         & ratio & Profiled with nvidia-smi\\
\gls{ram} used          & ratio & Profiled with the psutil Python library\\
Training duration       & ratio & Profiled with the datetime Python library\\
Number of \gls{flops}   & ratio & Measured once through the TensorFlow API\\
$F_1$ score             & ratio & Measured with the test set after the training phase\\
\rowcolor{gray!60}
\multicolumn{3}{l}{Other variables:} \\
\rowcolor{gray!20}
Dataset                 & nominal   & Three datasets: Chesslive, Caltech101, Stanford Dogs\\
\bottomrule
\end{tabular}%
\end{table}

\subsection{Datasets}\label{sec:dataset}
\subsubsection{Caltech101}
For the Caltech101 dataset~\cite{li_andreeto_ranzato_perona_2022}, we use the version by TensorFlow Datasets.\footnote{\url{https://www.tensorflow.org/datasets/catalog/caltech101}} The dataset consists of pictures of objects belonging to 101 classes, plus one background clutter class. Each image is labeled with a single object. Each class contains roughly 40 to 800 images, totaling around 9k images. Images are of variable sizes, with typical edge lengths of 200-300 pixels. We use the default train and test splits provided by the TensorFlow API. There are 9,144 images, out of which 3,060 are used for training and 6,084 for testing.

\subsubsection{Stanford Dogs}
For the Stanford Dogs dataset~\cite{KhoslaYaoJayadevaprakashFeiFei_FGVC2011}, we use the TensorFlow Datasets version.\footnote{\url{https://www.tensorflow.org/datasets/catalog/stanford_dogs}} The dataset contains images of 120 breeds of dogs from around the world. This dataset has been built using images and annotations from ImageNet for the task of fine-grained image categorization. There are 20,580 images, with a default partitioning of 12,000 for training and 8,580 for testing.

\subsubsection{Chesslive dataset}
Our research group has created the Chesslive dataset for multiple research purposes. The dataset contains images of chessboard squares, either empty or with a piece on them. The dataset has been built using three different data sources: 3D rendered images published by \citet{Georg2021}, a public dataset from Roboflow~\cite{ChessPiecesDataset}, and our own set of images collected from a real chess tournament. The latter two were added to include images of real pieces to get data similar to the one obtained from the setting where the model will be used. The pictures from the three sources are from full boards, not single pieces. Hence, we used the algorithm proposed by \citet{Georg2021} to crop the board squares from the images. After processing the data, the dataset consists of 34,010 50$\times$50 color images where 50\% belong to the positive class (occupied square) and 50\% to the negative class (empty square). As this is a custom dataset, we use a 70\%/30\% train/test split. Hence, there are 23,807 images for training and 10,203 for testing.

In all three cases, we do not set aside a portion of the dataset for validation because model optimization through hyperparameter tuning is beyond the scope of this analysis.

\subsection{Experiment setting}
To implement the model architectures, we utilize the implementations provided in TensorFlow\footnote{\url{https://www.tensorflow.org/versions/r2.10/api_docs}} for six base models: MobileNet V2, NASNet Mobile, ResNet-50, Xception, VGG-16, and Inception V3. Each model considerably differs in the number of parameters, FLOPs, and storage size (see Table~\ref{tab:models-meta-data}). We apply transfer learning to train our models. As such, we use the base models without their top, i.e., the fully-connected and output layers, and replace them with a new top. We follow the same training process for the first five models. We start by fixing the same hyperparameters for the five models, then we train each model on the Chesslive dataset in two consecutive steps: (i) train the models with all the layers frozen except for the top; and (ii) unfreeze all the layers and fine-tune the whole network. The Inception V3 model is trained on the Caltech101 and Stanford Dogs datasets. Its training process consists of a single step in which we attach a new top to the base model and train the model while the base model's layers are frozen. To speed up experimentation, the first five models are not utilized in \textit{\gls{rq}$_{3.1}$}. Also, the Inception V3 model is only used during \textit{\gls{rq}$_3$}.

\begin{table}[htb]
    \centering
    \caption{Number of parameters, GFLOPs, and size of each model architecture used in the study. All data have been extracted using an input size of 128x128.}\label{tab:models-meta-data}
    \begin{tabular}{cccc}
        \toprule
        \thead{Model architecture}  & \thead{Total parameters}    & \thead{GFLOPs}   & \thead{Size (MB)}\\ \midrule
        MobileNet V2                & 2,914,369                   & 0.20            & 11.12 \\
        NASNet Mobile               & 4,811,413                   & 0.37            & 18.35 \\
        ResNet-50                    & 26,211,201                  & 2.53            & 99.99 \\
        Xception                    & 22,960,681                  & 2.96            & 87.59 \\
        VGG-16                       & 14,715,201                  & 10.53           & 56.13 \\
        Inception V3                & 40,797,062                  & 1.50            & 155.63 \\
        \bottomrule
    \end{tabular}
\end{table}

Regarding the training environment, we use an Nvidia GTX 750 Ti-2GB in Desktop N and an Nvidia RTX 3070 8GB in Desktop \gls{ml}. Both settings use a 32 GB 3,600 MT/s DDR4 of memory. For the Server environment, we use the Universitat Polit{\`e}cnica de Catalunya's rdlab\footnote{\url{https://rdlab.cs.upc.edu/}} HPC service, including an Nvidia RTX 3090-24GB and 16 GB 2,666 MT/s DDR3 of memory for \textit{\gls{rq}$_1$} and \textit{\gls{rq}$_2$}, and 8 GB to 32 GB for \textit{\gls{rq}$_{3.1}$}.

We use nvidia-smi to collect the following \gls{gpu} metrics: power draw, utilization, and temperature. We use the psutil library to measure the amount of \gls{ram} used during the training process. We set the sampling interval to 1 second. The \gls{gpu} energy consumption is computed as the integral of the power over the training time. To compute the \gls{ram} energy consumption, we first transform the measurements of used \gls{ram} to power using the ratio 0.375W/GB of \gls{ram} used~\cite{bouzaHowEstimateCarbon2023}. Then we compute the \gls{ram} energy consumption as the integral of the power over training time. Finally, we sum the \gls{gpu} and \gls{ram} energy consumption to obtain the combined energy consumption.

The complete experiment and data analysis are implemented in Python 3.9.13.\footnote{\href{https://www.python.org/downloads/release/python-3913/}{Python 3.9.13 release}} The code used to compute \gls{flops} and analyze the data is available in our \textbf{replication package}~\cite{replicationPackage}.

\subsection{Experiment execution}\label{sec:execution-plan}
To answer \gls{rq}$_1$ and \gls{rq}$_2$, we define a 5x3 factorial design where the model architecture factor has five levels and the training environment factor has three levels (see Table~\ref{tab:variables}). For each $model \ architecture\times training \ environment$, we run the training process 30 times to provide more reliable measurements of the dependent variables. This design allows us to measure the energy consumption, \gls{gpu} usage, and model accuracy under different combinations of model architecture and training environment, allowing us to study their effects and interactions on all dependent variables. Additionally, conducting a factorial design provides us with all the required data to answer the \glspl{rq}. We only need to select the relevant factors and levels for each \gls{rq} (see Section~\ref{sec:data-analysis}).

To answer \gls{rq}$_3$, we first need to understand what parameters can influence energy consumption so we can use them to estimate it. This process corresponds to \gls{rq}$_{3.1}$, which we answer through two experiments using the Inception V3 model architecture.

The first experiment is run in the Desktop ML environment and employs a 3×3 factorial design with batch and input sizes as factors. Each of the nine combinations is trained 30 times while measuring the energy consumption. The experiment uses two datasets, Caltech101 and Stanford Dogs, to account for dataset-related variations. This design enables a robust analysis of batch and input sizes' main and interaction effects on energy consumption.

In the second experiment, we train the Inception V3 model on the Stanford Dogs dataset in the Server environment while changing the allocated \gls{ram} to 8 GB, 16 GB (default in previous experiments), 24 GB, and 32 GB. The batch size is fixed to 256, and the input size to 128x128. With this experiment, we want to verify if allocating additional \gls{ram} could help reduce energy consumption, suggesting it is a relevant factor to estimate the final consumption.

To ensure the correct measurement of energy consumption, we follow the guide by Cruz~\cite{Cruz2021}. We perform a dummy \gls{gpu}-intensive warm-up task before all the experiments, carried out by training one of the \gls{dl} models for approximately 10 minutes. Additionally, we schedule pauses of 5 minutes between the 30 runs of each experiment.

\gls{rq}$_{3.2}$ regards the study of patterns in energy consumption within the training process, and the development of an energy estimation method. To answer this \gls{rq}, we decided not to execute more experiments, but to use all the data we have already collected during the executions of all the experiments previously described.

In \gls{rq}$_{3.3}$ we evaluate the proposed energy estimation methods and compare their performance with widely used energy estimation methods (i.e., \gls{mlco2} and \gls{ga}).

\subsection{Data analysis}\label{sec:data-analysis}
This section outlines the preprocessing steps and statistical methods used to analyze the experimental data.

\subsubsection{Data processing}
We first cleaned the dataset by removing failed runs, empty values, and runs where the sampling rate exceeded the epoch duration, reducing the original 1,630 runs to 1,365 valid entries. Outliers were then removed separately for each dependent variable to preserve valid data for variables where the values were not extreme. Detailed steps are available in the replication package.

For \textit{\gls{rq}$_1$} and \textit{\gls{rq}$_2$}, we excluded runs from the Desktop N environment, where only MobileNet V2 and NASNet Mobile could be trained due to \gls{gpu} \gls{ram} constraints. These runs were retained to support pattern analysis in \textit{\gls{rq}$_3$}. Finally, we normalized total energy consumption by the number of images processed during training to enable fair comparisons across models and environments, accounting for differences in training time and model complexity, inspired by the approach of \citet{Fischer2023}.

\subsubsection{Statistical Analysis}

To address our three research questions, we applied regression models aligned with the characteristics of each outcome variable and verified their validity through residual diagnostics and model-specific transformations.

For \textit{\gls{rq}$_{1.1}$}, we fitted two linear models: one for normalized energy consumption and another for \gls{gpu} usage. Both models included architecture, training environment, and their interaction as predictors. Because \gls{gpu} usage is bounded between 0 and 100 and exhibited left-skewness, we applied a square-root transformation to the reflected values, using the formula $\sqrt{100 - \text{gpu\_usage}}$. 
Type III ANOVA with sum-to-zero contrasts was used to assess the significance of main effects and interactions, followed by Tukey-adjusted estimated marginal means for post-hoc comparisons.

To answer \textit{\gls{rq}$_{1.2}$}, we employed two linear models: one predicting normalized energy consumption and the other predicting average power draw. Both models included \gls{gpu} usage, model architecture, training environment, and their interactions. Model fit and statistical significance were assessed using the same ANOVA procedure above.

In \textit{\gls{rq}$_2$}, we examined the trade-off between energy consumption and model accuracy in two steps. First, $F_1$ scores---continuous and bounded within [0,1]---were modeled using a \gls{gamlss} with a Beta distribution, incorporating model computational complexity (\gls{flops}) as the main effect. The training environment showed no significant impact and was excluded from the final model. We assessed significance using Type III ANOVA and performed pairwise comparisons via Tukey-adjusted estimated marginal means. Following the evaluation of model complexity, we computed the Pareto front to identify optimal models that balance energy consumption and accuracy.

For \textit{\gls{rq}$_{3.1}$}, we fitted a \gls{glm} with a Gamma distribution to examine the impact of batch size, input size, and dataset on normalized energy consumption, including all interaction terms. Statistical significance was tested using Type III ANOVA. Additionally, a separate model assessed the influence of allocated RAM on energy usage, using a single predictor and evaluated via Type I ANOVA due to the absence of interactions. For \textit{\gls{rq}$_{3.2}$}, we examined power usage time series for recurring motifs using the FLUSS segmentation algorithm proposed by \citet{Gharghabi2017}, as implemented in the \texttt{stumpy} Python library \citep{law2019stumpy}. For \textit{\gls{rq}$_{3.3}$}, we train different regression models for two of the proposed energy estimation methods. To train and evaluate these models, we use a 5-fold cross-validation on 70\% of the data. We use the remaining 30\% to compare the accuracy of our methods with the two approaches mentioned before. We use the \gls{rmse} to evaluate the accuracy of all the energy estimation methods. In addition, we use R$^2$ and \gls{mae} in the cross-validation process to provide further insights into the accuracy of the regression models.

All statistical tests were conducted using a 95\% confidence level, and full modeling details, including preprocessing scripts and diagnostic outputs, are available in the replication package.

\section{Results}\label{sec:results}
This section presents the experimental findings addressing the \glspl{rq} guiding this study.

\subsection{\texorpdfstring{\glsentrytext{rq}\textsubscript{1}: Impact of \glsentrytext{dl} model architecture and training environment in energy consumption and \glsentrytext{gpu} usage}{RQ1: Impact of DL model architecture and training environment in energy consumption and GPU usage}}

\begin{figure}[tb]
    \centering
    \includegraphics{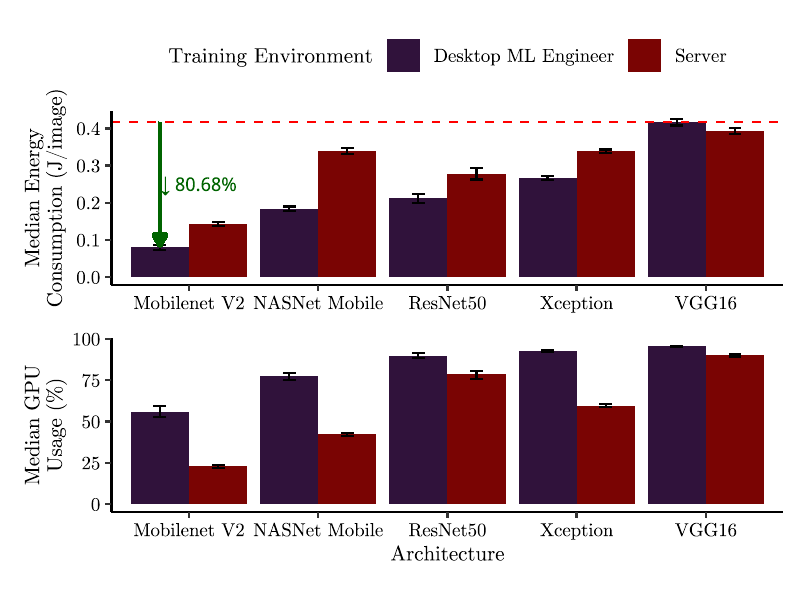}
    \caption[Comparison of the five model architectures trained in the different environments]{Comparison of the five model architectures trained in the different environments in terms of median energy consumption per image (top) and median \gls{gpu} usage (bottom). Architectures are ordered by ascending \gls{flops}.}
    \label{fig:environment-architecture-comparison}
\end{figure}

\textit{1) Results \gls{rq}$_{1.1}$}: As shown in Figure~\ref{fig:environment-architecture-comparison}, both energy consumption and \gls{gpu} usage during image classification training were significantly influenced by model architecture and training environment.

MobileNet V2 consistently showed the lowest energy consumption across both environments, while VGG-16 exhibited the highest. Although some trends held across environments, notable deviations emerged. In the Desktop ML environment, energy usage increased consistently with model computational complexity. However, this pattern broke in the Server environment, where ResNet-50 consumed less energy than NASNet Mobile, and NASNet Mobile and Xception exhibited comparable energy use.

Linear model analysis revealed significant main and interaction effects on per-image energy consumption ($p < .001$). Tukey-adjusted post hoc comparisons showed that MobileNet V2 trained in the Desktop ML environment was the most energy-efficient configuration, exhibiting significantly lower energy usage than all others (all $p < .0001$), achieving an 80.68\% reduction compared to the most demanding setup (VGG-16--Desktop ML). Architecture was the dominant explanatory factor ($\eta^2 = 0.851$), although it overlapped with the interaction effect ($\eta^2 = 0.065$) due to collinearity. Training environment explained much less variance on its own ($\eta^2 = 0.083$), but meaningfully modified architecture effects (especially for VGG16). Thus, its importance lies in its interaction with architecture rather than as a dominant main effect.

Interestingly, the Server environment did not consistently result in lower energy consumption. Except for VGG-16, all models consumed more energy when trained on the Server. For instance, NASNet Mobile’s energy consumption was significantly higher in the Server than in Desktop ML (estimate = --0.155, $p < .0001$), with similar trends for MobileNet V2, ResNet-50, and Xception. Additionally, the energy consumption of NASNet Mobile and Xception was statistically indistinguishable ($p = 1$), confirming visual observations. These results highlight that greater computational capacity does not uniformly lead to improved energy efficiency.

Regarding \gls{gpu} usage, \figurename~\ref{fig:environment-architecture-comparison} reveals substantial variation across model architecture--training environment combinations. VGG-16 trained in Desktop ML recorded the highest utilization, followed closely by Xception in the same setup. In contrast, MobileNet V2 in the Server environment showed the lowest usage, followed by NASNet Mobile.

The linear model fitted to reflected-transformed \gls{gpu} usage data revealed significant main and interaction effects ($p < 0.001$ for all). Architecture accounted for most of the variance ($\eta^2 = 0.679$), followed by environment ($\eta^2 = 0.269$) and the interaction ($\eta^2 = 0.051$). Partial $\eta^2$ values exceeded 0.98, indicating robust effects.

Tukey-adjusted pairwise comparisons showed that 44 of 45 architecture-environment pairs differed significantly ($p < .0001$), except for ResNet-50 in Desktop ML and VGG-16 in Server, which had similar \gls{gpu} usage.

\begin{tcolorbox}[colback=blue!5!white, colframe=blue!50!black, title=\faLightbulb~Answer to \gls{rq}$_{1.1}$]
    Model architecture and training environment have a substantial, non-uniform effect on energy consumption and \gls{gpu} usage during training. Lightweight models (e.g., MobileNet V2, NASNet Mobile) show high variability across environments, while heavier ones (e.g., VGG-16) consistently incur high computational costs. Efficiency depends on the model architecture--training environment combination. No environment is universally optimal, but in resource-constrained contexts, lightweight models in moderately provisioned environments provide the most favorable trade-offs.
\end{tcolorbox}

\begin{figure}[hbt]
    \centering
    \includegraphics{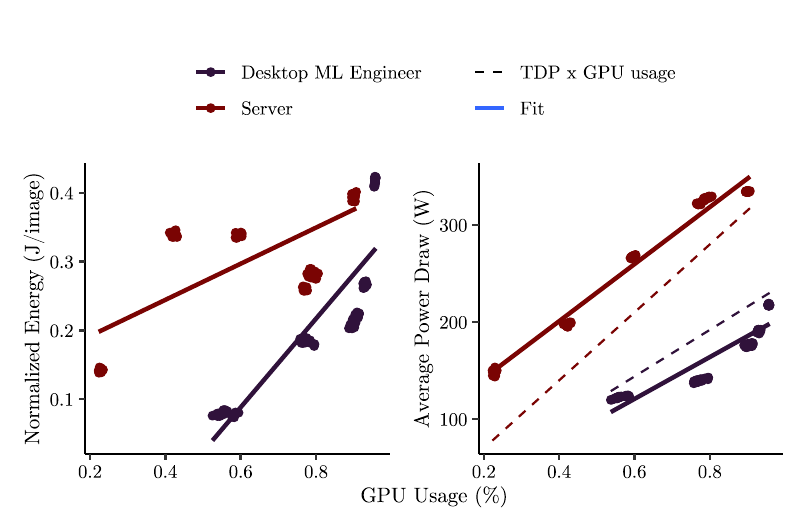}
    \caption{Relationships between \gls{gpu} usage, and normalized energy and power draw.}
    \label{fig:energy-and-power-vs-gpu}
\end{figure}

\textit{2) Results \gls{rq}$_{1.2}$}: \figurename~\ref{fig:energy-and-power-vs-gpu} illustrates the relationship between \gls{gpu} usage, normalized energy consumption (left panel), and average power draw (right panel). In the Desktop \gls{ml} environment, energy consumption scales linearly with GPU usage up to $\approx$80\%, after which it increases exponentially. In contrast, the Server environment shows no clear linear pattern, though a general positive trend is present. Notably, operating at 40\% GPU usage in the Server environment can result in similar---or even higher---energy consumption compared to 60\% or 80\% usage levels.

To further explore these relationships, we applied the linear model introduced in Section~\ref{sec:data-analysis}. Type III ANOVA revealed a significant main effect of model architecture \((F = 19.6, p < .001)\), along with significant two-way interactions between \gls{gpu} usage and architecture \((F=22.88, p < .001)\) and between model architecture and training environment \((F = 5.37, p<.001)\). A significant three-way interaction was also observed between \gls{gpu} usage, model architecture, and training environment \((F=4.58, p=.001)\).

Total eta-squared indicated that training environment and model architecture each explained approximately $\approx$37\% of the total energy consumption variance, while \gls{gpu} usage accounted for $\approx$22\%. Interaction effects contributed less than 3\% to the total variance.

The right panel of Figure~\ref{fig:energy-and-power-vs-gpu} depicts the relationship between \gls{gpu} usage and average power draw. In the Desktop \gls{ml} environment, the trend resembles that of energy consumption: a steep rise beyond 80\% usage, suggesting diminishing returns in energy efficiency. In the Server environment, power draw increases more linearly with \gls{gpu} usage. This divergence from the energy trend is likely due to differences in image processing time across model architectures, which affects total energy despite similar power levels.

Finally, we highlight a key limitation of commonly used power estimation practices. Estimating average power via \gls{tdp} scaled by \gls{gpu} utilization leads to substantial inaccuracies: it overestimates power in the Desktop \gls{ml} setup and underestimates it in the Server. This renders \gls{tdp}-based methods unreliable for cross-environment comparisons of model energy efficiency.

\begin{tcolorbox}[colback=blue!5!white, colframe=blue!50!black, title=\faLightbulb~Answer to \gls{rq}$_{1.2}$]
    The effect of \gls{gpu} usage on energy consumption is shaped by both training environment and model architecture. High \gls{gpu} usage can exponentially increase energy costs depending on the environment. TDP-based power estimation methods are inaccurate across environments, underscoring the need for more robust alternatives.
\end{tcolorbox}

\subsection{\texorpdfstring{\glsentrytext{rq}\textsubscript{2}: Trade-offs between energy consumption and model accuracy}{RQ2: Trade-offs between energy consumption and model accuracy}}

A preliminary analysis of descriptive statistics (Table~\ref{tab:models-f1}) revealed minimal differences in mean $F_1$ scores across model architectures, except for ResNet-50. The largest drop in performance—0.16—was observed between VGG-16 and ResNet-50. To assess the influence of computational complexity, we fitted a \gls{gamlss} model using GFLOPs to predict both the mean and dispersion of $F_1$ scores. The inclusion of GFLOPs led to a substantial model improvement over the null \(\Delta AIC \approx 1000\), with GFLOPs emerging as a highly significant predictor ($p < .001$). Tukey-adjusted pairwise comparisons confirmed significant differences across all GFLOPs levels ($p < .001$), with ResNet-50 (2.5 GFLOPs) performing significantly worse than both less and more complex models.

\begin{table}[htb]
    \centering
    \caption{Descriptive statistics for $F_1$ score by model architecture.}\label{tab:models-f1}
    \begin{tabular}{ccccc}
        \toprule
                                    &                 & \multicolumn{3}{c}{\thead{$F_1$ score}} \\ \Xcline{3-5}{1pt}
        \multirow{-2}{*}{\thead{Model architecture}}  & \multirow{-2}{*}{\thead{GFLOPs}}  & \thead{Mean}    &\thead{Std.}   &\thead{Median}\\ \midrule
        MobileNet V2                & 0.20            & 0.97    & 0.015     & 0.97\\
        NASNet Mobile               & 0.37            & 0.98    & 0.001     & 0.98\\
        ResNet-50                    & 2.53            & 0.83    & 0.168     & 0.93\\
        Xception                    & 2.96            & 0.98    & 0.001     & 0.98\\
        VGG-16                       & 10.53           & 0.99    & 0.001     & 0.99\\
        \bottomrule
    \end{tabular}
\end{table}

\begin{figure}[ht]
    \centering
    \includegraphics{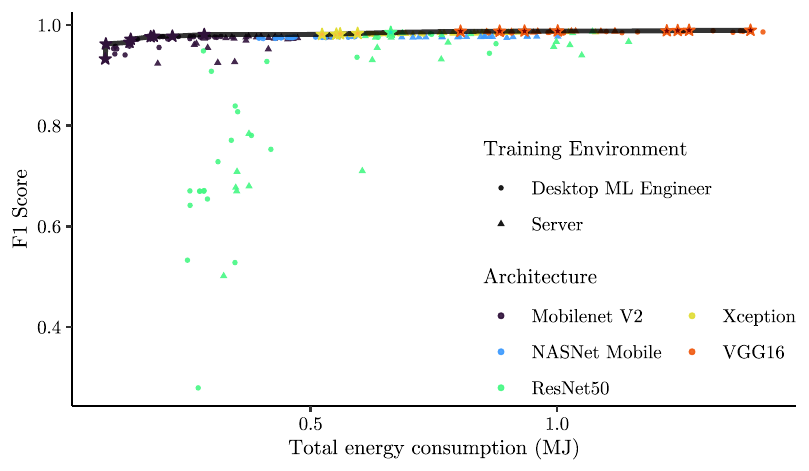}
    \caption{Distribution of $F_1$ score vs. energy consumption. Instances belonging to the Pareto-optimal configurations are represented with a star.}
    \label{fig:pareto-front-f1-vs-energy}
\end{figure}

Having established a significant effect of model architecture on both energy consumption and $F_1$ score, we examined their trade-offs. \figurename~\ref{fig:pareto-front-f1-vs-energy} displays the total energy consumption versus $F_1$ score across all training runs, with the Pareto front highlighted. Inspection of this front revealed that NASNet Mobile does not contribute any optimal instances, suggesting that other models—especially MobileNet V2—offer more favorable trade-offs. Indeed, nearly half of the Pareto-optimal configurations correspond to MobileNet V2.

\begin{tcolorbox}[colback=blue!5!white, colframe=blue!50!black, title=\faLightbulb~Answer to \gls{rq}$_2$]
    Our analysis shows that energy-intensive models yield negligible gains in $F_1$ score compared to energy-efficient alternatives in our context. For example, replacing VGG-16 with MobileNet V2 reduces energy consumption by approximately 80\%, with only a 0.01 drop in $F_1$ score.
\end{tcolorbox}

\subsection{\texorpdfstring{\glsentrytext{rq}\textsubscript{3}: Energy estimation model}{RQ3: Energy estimation model}}

\begin{figure}[hbt]
    \centering
    \includegraphics{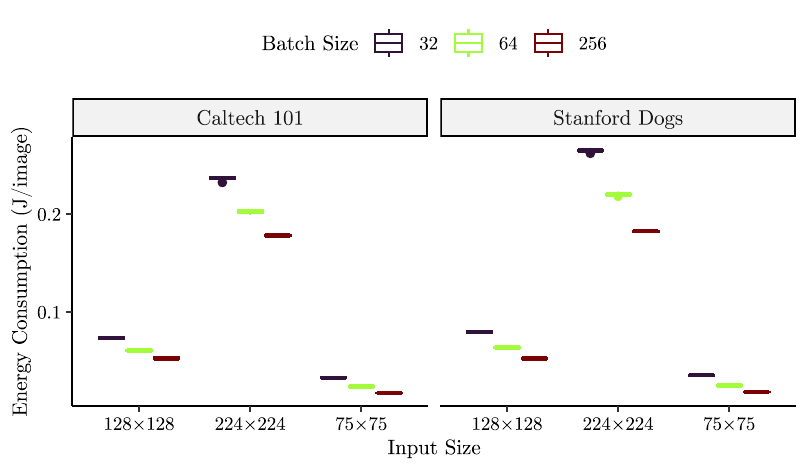}
    \caption[Impact of batch and input size on energy consumption by dataset]{Impact of batch and input size on normalized energy consumption for Inception V3 in the Desktop ML environment, split by dataset.}
    \label{fig:batch-input-size-impact}
\end{figure}

\textit{1) Results \gls{rq}$_{3.1}$: Identifying useful parameters for estimating energy consumption.} To identify useful predictors of energy consumption during training, we analyzed the effects of batch size and input size. \figurename~\ref{fig:batch-input-size-impact} presents normalized energy consumption across varying configurations for the Inception V3 model, trained on the Desktop ML environment using the Caltech101 and Stanford Dogs datasets.

\begin{figure}[tb]
    \centering
    \includegraphics[width=\textwidth]{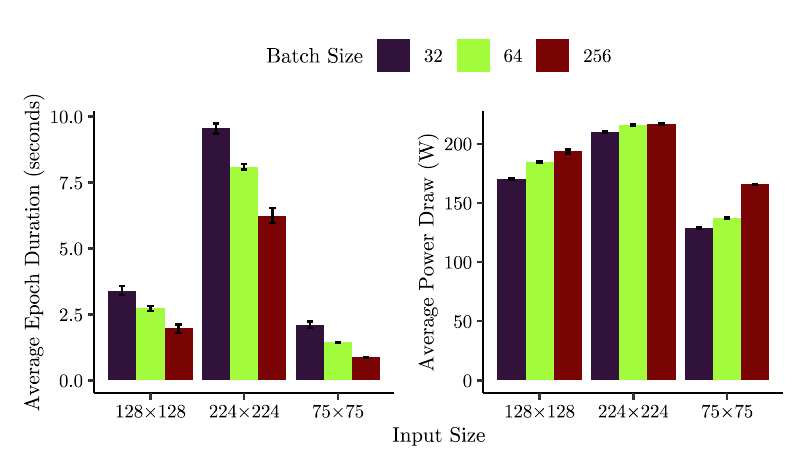}
    \caption{Impact of batch and input size on average epoch duration and average power consumption for the Desktop ML training environment and Caltech 101 dataset.}
    \label{fig:batch-input-size-impact-on-time-and-power}
\end{figure}

Batch size demonstrated a consistent energy-saving effect: larger batches significantly reduced energy consumption per image across all input sizes and datasets. This reduction is largely attributed to decreased training time, despite a modest increase in average power usage, as illustrated in \figurename~\ref{fig:batch-input-size-impact-on-time-and-power}.

In contrast, increasing input size substantially elevated energy consumption. For example, input dimensions of 224$\times$224 resulted in nearly four times higher energy usage compared to 75$\times$75. This increase is driven by the associated rise in computational complexity (\gls{flops}), reflected in both longer training durations and greater power draw, regardless of batch size or dataset.

The \gls{glm} results confirmed that batch size, input size, dataset, and their interactions were all statistically significant ($p < .001$). Effect size analysis revealed that input size was the dominant predictor, accounting for 94.7\% of the variance in energy consumption ($\eta^2 = 0.947$). Batch size explained 3.4\% ($\eta^2 = 0.034$), while dataset effects were negligible ($\eta^2 = 0.002$). Although absolute interaction effects were small (e.g., $\eta^2 = 0.014$ for batch size $\times$ input size), partial effect sizes were large, indicating strong contextual influence (e.g., partial $\eta^2 = 0.993$).

Tukey-adjusted pairwise comparisons by dataset confirmed significant differences between batch–input combinations ($p < 0.001$). Hedges’ \textit{g} analysis further emphasized the magnitude of these effects. Input size increases led to very large effect sizes—exceeding $g = 200$ when comparing 75$\times$75 to 224$\times$224. For example, training with batch size 32 and input size 75$\times$75 consumed substantially less energy than with 224$\times$224 inputs: $g = -214.2$ (Caltech101) and $g = -218.4$ (Stanford Dogs). Even moderate increases (e.g., to 128$\times$128) produced large effects ($g = -87.5$ to $-130.6$).

Larger batch sizes consistently reduced energy usage, particularly at lower input sizes. For example, increasing the batch size from 32 to 256 at 75$\times$75 size yielded large positive effects ($g = 71.1$ for Caltech101, $g = 70.9$ for Stanford Dogs), indicating substantial energy savings per image. In contrast, increasing batch size at high input sizes (e.g., 224$\times$224) had smaller but still notable effects (e.g., $g = 30.6$ for Caltech101 and $g = 40.3$ for Stanford Dogs when comparing batch sizes 32 and 256).

The most substantial energy reductions were observed when the batch size increased and the input size decreased. These consistent, large effect sizes across datasets highlight the strong practical impact of these training parameters on energy efficiency in \gls{dl}.

\begin{figure}[tb]
   \centering
   \includegraphics{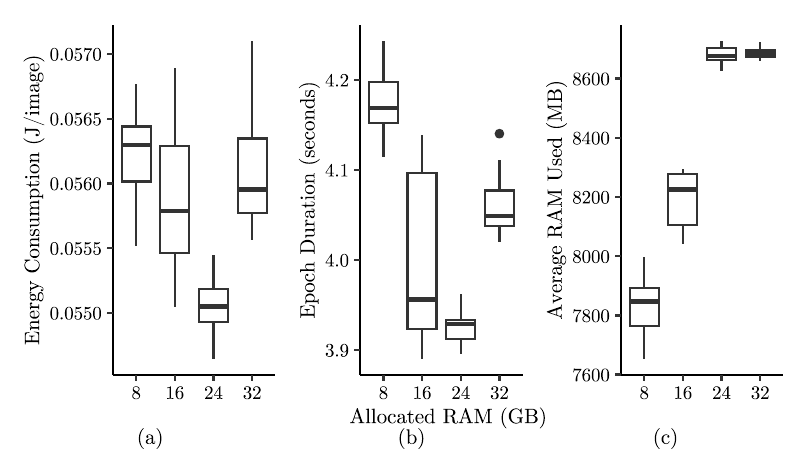}
   \caption{Effect of allocated \gls{ram} in; (a) normalized energy consumed, (b) epoch duration, and (c) average \gls{ram} used in the Server environment.}
    \label{fig:ram-impact}
\end{figure}

Moving to the effect of allocated \gls{ram}, \figurename~\ref{fig:ram-impact} presents boxplots of normalized training energy consumption (J/image), epoch duration (seconds), and average \gls{ram} usage (GB), grouped by total allocated \gls{ram} (GB). The visual trends indicate a moderate reduction in both energy consumption and epoch time as allocated RAM increases, although the improvements diminish beyond 24GB.

Notably, performance gains are non-linear. Increasing allocation beyond a certain point can be negligible or even counterproductive, as shown by the overlapping distributions across groups. For example, the energy consumption distributions for 8GB, 16GB, and 32GB overlap substantially. Similarly, epoch duration for 16GB closely overlaps with that of 24GB and 32GB. In terms of average RAM used, the 24GB and 32GB groups are nearly identical.

Statistical analysis supports these visual observations. Allocated RAM had a significant effect on energy consumption per image ($F = 59.41$, $p < .001$), explaining 63\% of the variance ($\eta^2 = 0.63$). Tukey-adjusted comparisons revealed a significant reduction in energy consumption from 8GB to 24GB ($p < .001$), but no significant benefit from further increasing to 32GB ($p = .30$).

Epoch duration was also significantly influenced by allocated RAM ($F = 144.98$, $p < .001$), accounting for 81\% of the variance ($\eta^2 = 0.81$). While increasing from 8GB to 24GB led to significant speedups ($p < .001$), a further increase to 32~GB slightly increased duration, albeit still statistically significant ($p < .001$).

As expected, allocated RAM strongly affected average RAM usage during training ($F = 1087.59$, $p < .001$, $\eta^2 = 0.97$). However, RAM usage plateaued between 24GB and 32GB, with no statistically significant difference between them ($p = .99$).

\begin{tcolorbox}[colback=blue!5!white, colframe=blue!50!black, title=\faLightbulb~Answer to \gls{rq}$_{3.1}$]
Batch size, input size, and allocated \gls{ram} are all significant predictors of energy consumption during training. Larger batch sizes consistently reduce the energy per image by decreasing training time, while larger input sizes significantly increase energy consumption due to higher computational demands. Allocated \gls{ram} also influences energy use, with improvements observed up to a threshold. Beyond this point, additional RAM yields diminishing or even adverse effects. Among all parameters, input size is the most dominant factor for energy modeling, followed by batch size and allocated RAM.
\end{tcolorbox}

\textit{2) Results \gls{rq}$_{3.2}$: Predicting training energy consumption.} To develop effective models for predicting energy consumption during training, we first investigate temporal and power-related stability within training runs.

We begin by assessing whether epoch duration is consistent across executions. As shown in \figurename~\ref{fig:batch-input-size-impact-on-time-and-power}, the mean and standard deviation of epoch durations for the Inception V3 model on Caltech101 indicate very low variability. Narrow error bars across batch and input size combinations suggest strong temporal stability. This observation generalizes across datasets, models, and environments, indicating that epoch duration remains highly consistent under fixed conditions.

\begin{figure}[tb]
    \includegraphics[width=\textwidth]{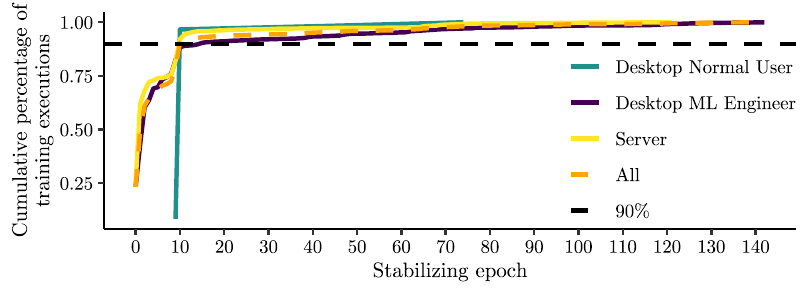}
    \caption{Cumulative percentage of training executions that stabilize after the n-th epoch in the different training environments.}\label{fig:stabilizin-epoch-distribution}
\end{figure}

Next, we explore power usage patterns across training runs. Assuming that power stabilizes after an initial warm-up period, we apply semantic segmentation to detect when steady-state behavior emerges. \figurename~\ref{fig:stabilizin-epoch-distribution} reveals that in at least 90\% of the runs, power consumption stabilizes by epoch ten. Thus, we define epoch ten as the ``stabilizing epoch'' for subsequent analysis.

Given the stability of both epoch duration and power consumption, we propose the Stable Training Epoch Projection (STEP) methods:

1. Stable Training Epoch Power Projection (STEP-P):

\begin{equation}
    E = \overline{P}_w \times t;\ 1 \le w \le s\label{eq:method-1}
\end{equation}

\noindent where $\overline{P}_w$ is the average power over a window of $w$ stable epochs, $t$ is the total training time, and $s$ is the total number of stable epochs.

2. Stable Training Epoch Energy Projection (STEP-E):

\begin{equation}
        E = E_w \times \dfrac{s}{w} + E_u;\ 1 \le w\le s\label{eq:method-2}
\end{equation}

\noindent where $\overline{E}_w$ is the total energy of a window of $w$ stable epochs, and $E_u$ is the energy from unstable epochs.

\begin{figure}[ht]
    \centering
    \includegraphics[width=\textwidth]{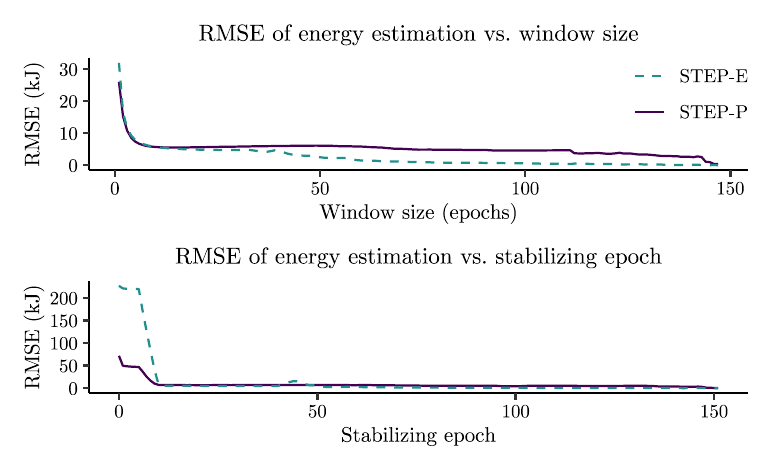}
    \caption{Effect of changing $w$ and the stabilizing epoch on the \gls{rmse} of the STEP methods. The top chart shows the progression of the \gls{rmse} when increasing $w$ from the stabilizing epoch. The bottom chart shows the same progression when using $w=5$ and changing the stabilizing epoch.}
    \label{fig:energy-estimation-vs-window-size-and-stabilizing-epoch}
\end{figure}

\figurename~\ref{fig:energy-estimation-vs-window-size-and-stabilizing-epoch} (top) shows the impact of window size $w$ on estimation accuracy, measured using RMSE (in kilojoules). A window size of five already achieves strong performance, with RMSE values of 7.34 for STEP-P and 7.88 for STEP-E.

We also examine the influence of the stabilizing epoch (bottom chart in \figurename~\ref{fig:energy-estimation-vs-window-size-and-stabilizing-epoch}). Results confirm that both methods reach low and stable RMSE values after epoch ten. This aligns with the previously identified stabilization point.

Finally, using a randomly selected window after the stabilizing epoch (rather than starting precisely at epoch ten) slightly improves estimation accuracy: RMSE drops to 6.96 for STEP-P and 5.13 for STEP-E---representing modest improvements of 0.34 and 2.75 kJ, respectively.

While the STEP methods demonstrate strong predictive accuracy, they rely on runtime measurements and are therefore unsuitable for estimating energy consumption before training. To address this limitation, we introduce the Pre-training Regression-based Estimation (PRE) methods, which do not require in-process measurements and can be used in advance. One predicts the average power draw over the entire training execution (PRE-P), while the other estimates the average energy consumption per stable epoch (PRE-E).

These models take as input a set of parameters typically available to practitioners before training begins. These include the number of CUDA cores on the \gls{gpu}, total \gls{gpu} \gls{ram}, the \gls{gpu}'s thermal design power (\gls{tdp}), total system \gls{ram}, the model's GFLOPs, the number of training and validation samples, image width and height, and the batch size.

This approach differs from existing estimation tools such as \gls{mlco2}, which estimates energy as:

\begin{equation}
    E = t \times P_g
\end{equation}

where $t$ is the runtime in hours and $P_g$ is the \gls{gpu}'s \gls{tdp}. Similarly, \gls{ga} estimates energy as:

\begin{equation}
    E = t \times (n_c \times P_c \times u_c + n_g \times P_g \times u_g + n_m \times P_m) \times PUE \times k
\end{equation}

where $n_c$, $n_g$, and $n_m$ are the counts of CPU cores, GPU cores, and available memory (GB), respectively; $P_c$ and $P_g$ are the TDPs of the CPU and GPU; $u_c$ and $u_g$ are their usage factors (between 0 and 1); $P_m$ is RAM power draw, set to 0.3725 W per GB; $PUE$ is the power usage effectiveness of the data center; and $k$ accounts for repeated runs.

Tables~\ref{tab:power-regression-results} and \ref{tab:energy-regression-results} summarize the performance of four regression models—Linear, Ridge, Kernel Ridge, and Support Vector Machines (SVM)---evaluated using 5-fold cross-validation on 70\% of the dataset. For PRE-E, we consider two variants: one that includes the energy from initial (unstable) epochs, and one that excludes it. All models perform well; however, Kernel Ridge regression consistently yields the best results across all evaluation metrics. Accordingly, we select it as our final model. For PRE-E, we opt for the variant trained with initial epochs included, as it achieves slightly superior performance in two out of three metrics.

\begin{table}[htb]
    \centering
    \begin{minipage}{0.4\textwidth}%
        \centering
        \caption{Performance of the PRE-P models in the 5-fold cross-validation.}%
        \label{tab:power-regression-results}%
        \begin{adjustbox}{max width=\textwidth}%
        \begin{tabular}{lccc}
            \toprule
             Regression method & R$^2$ & \gls{rmse} & MAE \\
            \midrule
            Linear & 0.83 & 28.54 & 21.25 \\
            Ridge & 0.82 & 28.60 & 21.27 \\
            \textbf{Kernel Ridge} & \textbf{0.99} & \textbf{7.85} & \textbf{4.38} \\
            SVM & 0.96 & 13.21 & 6.03 \\
            \bottomrule
        \end{tabular}%
        \end{adjustbox}
    \end{minipage}%
    \qquad
    \begin{minipage}{0.45\textwidth}%
            \centering
            \caption{Performance of the PRE-E models in the 5-fold cross-validation.}%
            \label{tab:energy-regression-results}%
            \begin{adjustbox}{max width=\textwidth}
            \begin{tabular}{lccccc}
            \toprule
            Regression method & R$^2$ & \gls{rmse} & MAE & \thead{Uses \\ initial epochs?} \\
            \midrule
            Linear                  & 0.83          & 1.62          & 0.97 & True \\
            Ridge                   & 0.84          & 1.60          & 0.95 & True \\
            \textbf{Kernel Ridge}   & 0.96          & \textbf{0.74} & \textbf{0.31} & True \\
            SVM                     & 0.95          & 0.86          & 0.38 & True \\
            Linear                  & 0.84          & 1.76          & 1.03 & False \\
            Ridge                   & 0.84          & 1.74          & 1.02 & False \\
            Kernel Ridge            & \textbf{0.98} & 0.78          & 0.33 & False \\
            SVM                     & 0.95          & 0.93          & 0.40 & False \\
            \bottomrule
            \end{tabular}%
            \end{adjustbox}
    \end{minipage}
\end{table}

The proposed PRE methods need an additional step to make their predictions actionable. For PRE-P, the predicted average power must be multiplied by the expected total training duration. Likewise, the output of PRE-E must be multiplied by the anticipated number of training epochs. This step translates the per-unit predictions into complete energy consumption estimates for the full training process.

\begin{tcolorbox}[colback=blue!5!white, colframe=blue!50!black, title=\faLightbulb~Answer to \gls{rq}$_{3.2}$]
    Training epochs exhibit highly stable behavior in both duration and power under fixed settings. We identify the 10th epoch as a reliable stabilization point and introduce the STEP methods, which can accurately predict total training energy using data from only a few stable epochs. Additionally, we propose the PRE methods, based on regression models, which estimate energy consumption before training using easily accessible hardware and model parameters.
\end{tcolorbox} 

\textit{3) Results \gls{rq}$_{3.3}$: Evaluation of the estimation methods.}

To evaluate the quality of the online and offline estimation methods, we test them on 30\% of the data reserved for evaluation, and compare them to \gls{mlco2} and \gls{ga}.

The STEP methods yield the lowest RMSE, with STEP-P reporting 5.50 kJ and STEP-E 7.97 kJ. PRE methods follow with PRE-P reporting an RMSE of 30.36 kJ, and PRE-P 58.98 kJ. The two methods already proposed in the literature obtain an \gls{rmse} of 159.83 kJ (\gls{ga}) and 185.39 kJ (\gls{mlco2}). These two methods more than duplicate the \gls{rmse} of the worst-performing of our proposed methods.

\begin{figure}[!ht]
    \includegraphics[width=\textwidth]{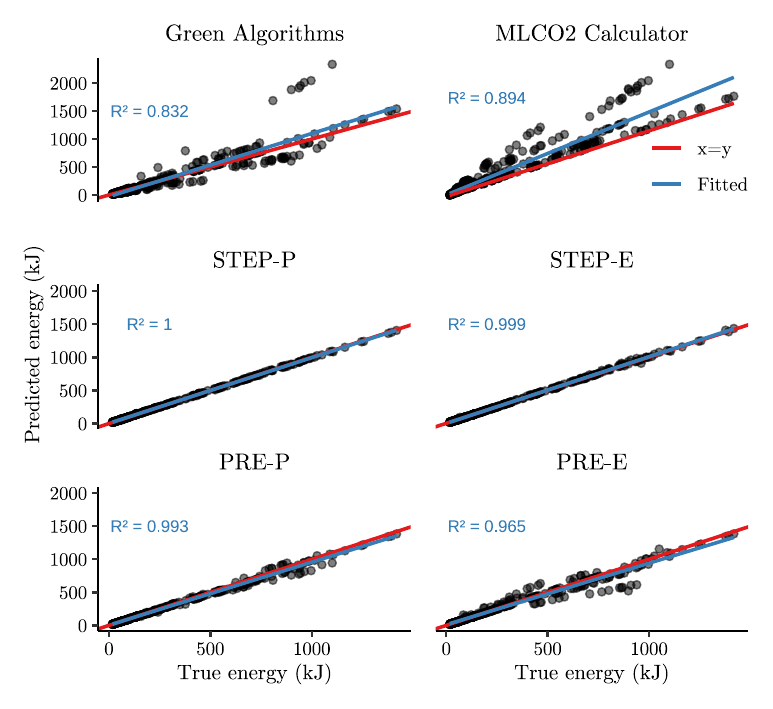}
    \caption{Predicted energy vs. real energy using six different energy prediction methods. The blue line shows the fitted line of the estimator. The red line is the perfect fit for the true energy.}
    \label{fig:estimation-methods-comparison}
\end{figure}

To further analyze the performance of the different methods, we compare the predicted energy vs. the true energy. In Figure~\ref{fig:estimation-methods-comparison}, we see how the \gls{ga}, \gls{mlco2}, and PRE-E have a noticeable spread from the ideal output. \gls{mlco2} is biased to overestimate energy consumption consistently. Similarly, \gls{ga} is slightly skewed to report overestimations. However, we see that it frequently reports underestimations as well. Instead, PRE-E is slightly skewed to underestimate energy consumption. Nevertheless, it also reports overestimations that drive far from the real energy.

Regarding the three remaining methods, we see how they report energy consumption with minimal deviations from the actual values. Indeed, the STEP methods almost fit the ideal values perfectly. This is no surprise since they benefit from estimating the energy from the actual measurements obtained while training.

\begin{tcolorbox}[colback=blue!5!white, colframe=blue!50!black, title=\faLightbulb~Answer to \gls{rq}$_{3.3}$]
    The proposed methods achieve two to three times lower error values, even with the worst-performing of the proposed approaches. Moreover, both STEP and PRE methods provide fairly accurate estimations, providing data scientists with accessible methods to obtain reliable energy estimations both before and during training.
\end{tcolorbox}

\section{Discussion}\label{sec:discussion}

\subsection{Model architecture and training environment: fiends or foes?}
Our findings highlight the critical influence of both model architecture and training environment on the energy consumption of training \gls{dl} models. Choosing an appropriate combination of architecture and environment can yield substantial energy savings---up to 80\%---with only a minor performance trade-off (e.g., a $0.01$ decrease in $F_1$ score when switching from VGG-16 to MobileNet V2 in the Desktop ML environment).

Importantly, we observe that energy efficiency is not solely a property of either the model or the environment; it results from their interaction. In particular, we find that extremely high or low \gls{gpu} utilization leads to inefficient energy use. At very high utilization (e.g., VGG-16 on Desktop ML), the \gls{gpu} is overburdened, leading to increased power consumption. Conversely, under-utilization (as seen in Server training of smaller models) wastes energy by powering unused hardware components.

This behavior appears to stem from a mismatch between the model's computational demands and the hardware's capabilities. Our results suggest that a \gls{gpu} will activate only as many cores as required by the model, and any surplus compute resources remain idle but still consume power. If a model requires $N$ cores and the \gls{gpu} provides $N+k$ (with $k \ll N$), the excess does not improve performance but increases energy usage unnecessarily. This observation aligns with the interaction effects reported in our statistical analysis and resonates with findings from prior work such as \citet{husomEngineeringCarbonEmissionaware2024}.

Moreover, although one might assume that more powerful environments, such as the Server environment, are inherently more energy-efficient, our results show this is only true in specific cases (e.g., VGG-16). Across most configurations, the Server did not yield faster training and often led to higher energy costs, challenging common assumptions about compute scaling.

These insights underline the need for collaboration between data scientists and software engineers when designing \gls{dl} training pipelines. Data scientists may select models for their accuracy or transferability, while engineers can ensure that computational resources match model requirements. This multidisciplinary cooperation is especially important when considering green AI practices.

Following the line of thought of \citet{abad_et_al:DagRep.11.4.34}, we suggest that serverless or adaptive computing infrastructures may offer energy-efficient alternatives. These systems can dynamically allocate hardware based on a model’s real-time requirements, reducing reliance on manual configuration and domain expertise.

Our results align with the findings of \citet{Georgiou2022} where they observed different energy and performance costs for different \gls{dl} frameworks. Our results also show that allocating more \gls{ram} does not guarantee improvements in training speed or energy efficiency. Over-allocation wastes energy and resources---especially in shared Server environments, where unused RAM could benefit other users. However, determining the precise RAM requirement for a model in advance remains a challenge. There is a clear need for intelligent tooling to guide users in selecting optimal resource configurations.

Ultimately, the ability to model and understand the relationship between training environments and model architectures has significant implications for reducing carbon emissions and optimizing resource usage. This calls for the development of standardized AI energy benchmarks, as also proposed by \citet{shiEfficientGreenLarge2025}. Such benchmarks would enable consistent evaluation of \gls{dl} models' energy profiles across environments and assist practitioners in making informed deployment decisions.

Efforts like the reporting guidelines from \citet{Castano2023} for Hugging Face, which include carbon impact metrics, are steps in the right direction. Future pre-trained models should recommend ideal hardware configurations, and \gls{dl} frameworks should go further, integrating energy monitoring and transparently reporting actual resource utilization.

\begin{tcolorbox}[colback=yellow!5!white, colframe=orange!70!black, title=\faCompass~Take-away 1]
    The most energy-efficient training environment is one that closely matches the computational requirements of the model. To reduce the carbon footprint of \gls{dl} projects, training environments should be carefully selected or dynamically adapted to fit the resource profile of the model being trained.
\end{tcolorbox}

\subsection{Server vs. Desktop, which environment to choose?}
While Desktop environments require the upfront purchase of a \gls{gpu}, they can offer substantial energy savings, provided subsequent model architectures are carefully chosen to match the hardware. Our results show that, for certain models, Desktop environments consistently consume less energy than Server environments. This reduced energy footprint can translate into lower operational costs, particularly when electricity prices are high. When accounting for the long-term cost of ownership---including the avoided expenses of server maintenance, infrastructure, and storage---Desktop environments can become economically competitive, with the added benefit of promoting accessibility and resource efficiency for smaller organizations and individual researchers.

However, the energy efficiency of training depends significantly on the alignment between 
models computational requirements and the available hardware. For large, complex models or workflows that demand substantial computational flexibility, Server environments offer a clear advantage. These environments typically support a wider range of \glspl{gpu} and configurations, making it easier to match model complexity to available compute power without requiring new hardware purchases. This adaptability is especially valuable in organizational settings where computational workloads vary across projects.

From an environmental perspective, while Desktop setups can offer lower operational energy consumption, one must also consider the embodied carbon footprint of \glspl{gpu}. Server environments can mitigate this impact by supporting shared and sustained usage of hardware resources, thereby improving the overall utilization of energy-intensive components. Moreover, server-class infrastructures can often be hosted in geographically optimized data centers powered by cleaner energy sources~\cite{patterson_carbon_2022}, a flexibility not achievable with local Desktop systems.

\begin{tcolorbox}[colback=yellow!5!white, colframe=orange!70!black, title=\faCompass~Take-away 2]
The decision between Server and Desktop should be guided by model complexity, training frequency, and resource amortization, both economic and environmental. The most sustainable strategy may involve a hybrid approach that combines the agility of Desktop environments for lightweight, frequent training tasks with the scalability and efficiency of Server infrastructures for heavier workloads.
\end{tcolorbox}

\subsection{Are we using the correct estimators for energy consumption?}
Our study reveals that widely adopted estimation methods and assumptions often fail to deliver accurate predictions of energy consumption in \gls{dl}. One such assumption is that the number of \gls{flops} required by a model is a reliable proxy for its energy usage. While increased \gls{flops} generally correlate with higher energy demands, our results reveal substantial variation across training environments. For instance, although Xception requires nearly eight times more \gls{flops} than NASNet Mobile, both consume almost the same energy in the Server environment---despite Xception being more energy-intensive in the Desktop \gls{ml} environment. This discrepancy highlights a key limitation of \gls{flops} as a standalone estimator: they do not account for training environment characteristics.

These findings align with prior critiques in the literature. \citet{Schwartz2020} and \citet{lacoste2019quantifying} suggest \gls{flops} as a proxy for energy, while others caution against hardware-agnostic metrics \cite{cao2020accurate, hendersonSystematicReportingEnergy2020, aspertiSurveyVariationalAutoencoders2021}. Our results reinforce the latter view, emphasizing the tight and complex interplay between model architecture, training environment, and energy behavior.

Another common estimation method assumes the \gls{tdp} of a \gls{gpu} approximates its average power draw during training~\cite{lacoste2019quantifying, lannelongueGreenAlgorithmsQuantifying2021, luccioni2023counting}. Although we find this method tends to overestimate energy use, making it conservative, it still introduces significant uncertainty. Specifically, it hampers fair comparisons across models, as one could easily mistake an overestimated energy footprint for a more energy-intensive model, when in reality, it may be more efficient.

This problem intensifies when \gls{tdp} values are scaled by \gls{gpu} utilization---an increasingly common practice aimed at incorporating GPU load into estimations. While this approach intends to improve accuracy, it can be more problematic than using unscaled \gls{tdp}, as it introduces the risk of both over- and underestimation. Our analysis shows that power consumption does not consistently scale linearly with \gls{gpu} usage, and the nature of this relationship varies across environments. For example, scaling \gls{tdp} by \gls{gpu} usage overestimated power in the Desktop \gls{ml} environment, but underestimated it in the Server environment. These inconsistencies highlight the unreliability of \gls{tdp}-based methods for estimating energy consumption, especially in cross-system comparisons.

These limitations have significant implications for evaluating the environmental sustainability of \gls{dl} models. Platforms such as Hugging Face encourage researchers to report \emissions{} and recommend tools like CodeCarbon or \gls{mlco2}. While real-time estimators, such as CodeCarbon, can provide accurate measurements, their adoption often requires modifying training scripts or relying on specific platforms (e.g., AutoTrain), barriers that many practitioners do not overcome. Moreover, practitioners may also wish to report the \emissions{} of older models for which no real-time measurements exist. In practice, this leads many to rely on offline tools such as \gls{mlco2}, which, as our results demonstrate, frequently misestimate energy use and distort comparisons.

This challenge is evident in the Hugging Face model repository: as of March 2023, fewer than 1\% of the 170,000 hosted models had reported emissions~\cite{Castano2023}. If creators were to update their models retroactively, roughly 168,000 would depend on tools such as \gls{mlco2} or \gls{ga}---underscoring the risk of widespread reliance on biased methods.

Instead, we advocate for a data-driven approach. Our findings indicate that such methods offer more accurate predictions, capturing the complex interactions between model, environment, and energy use. This perspective aligns with broader software engineering efforts to identify robust proxies for energy consumption~\cite{weber_twins_nodate}, a challenge that remains largely unresolved.

\begin{tcolorbox}[colback=yellow!5!white, colframe=orange!70!black, title=\faCompass~Take-away 3]
    Estimating energy consumption using \gls{flops} or scaled \gls{tdp} is unreliable, as these methods ignore the complex interplay between model architecture and training environment. Our findings show that accurate estimation requires data-driven, context-aware approaches that reflect both system and model characteristics.
\end{tcolorbox}

\subsection{Data-driven energy estimation}
In this study, we introduced several data-driven methods for estimating energy consumption that can improve the reliability and reproducibility of future work in Green IN \gls{ai}.

We first proposed the STEP methods that require only a fraction of the total training process to yield accurate results. While one might argue that it is preferable to measure the entire training using real-time energy estimators such as pynvml, RAPL, or CodeCarbon---especially since these tools impose minimal overhead---we believe the value of our STEP methods lies elsewhere. Specifically, they are particularly useful in cases where retraining is infeasible, such as with large-scale models like GPT-4 or BLOOM. For instance, BLOOM's training spanned over 1.08 million GPU hours, making a second run purely for energy estimation impractical. Yet, as \citet{luccioniEstimatingCarbonFootprint2023} demonstrated, such models are often evaluated using the inaccurate \gls{tdp}-based assumptions. Instead, using STEP-P, one can estimate energy consumption by measuring just a small, representative segment of the training process, offering a much more reliable alternative.

Following the idea of \citet{regueroEnergyefficientNeuralNetwork2025}, an interesting extension of the STEP approach is energy-aware early stopping. Much like data scientists monitor training loss or accuracy to determine when to halt training, energy consumption could be incorporated as a real-time constraint. By observing both energy usage and evaluation metrics during initial epochs, one could estimate the projected energy cost of continuing training relative to the model's expected improvement. This opens the door to developing smarter, energy-aware training pipelines.

Inspired by the STEP methods, we proposed the PRE methods. While slightly less accurate, these methods offer practical advantages: they require no model retraining and only depend on a small set of easily obtainable parameters (e.g., model architecture, batch size, input resolution, \gls{gpu} type). Compared to methods like \gls{mlco2}, which rely on coarse assumptions, our PRE methods provide a more reliable, context-sensitive estimation with minimal user effort.

PRE methods could contribute meaningfully to initiatives like Hugging Face's AI Energy Score~\cite{luccioniAIEnergyScore2025}. At present, the score focuses exclusively on inference energy due to the lack of standardized, verifiable training consumption data. Our methods could help close this gap, offering a common framework to estimate training energy based on transparent, repeatable inputs---without requiring developers to share proprietary logs or rerun models.

However, a key limitation of the PRE approach is generalizability, which depends heavily on the availability of diverse, high-quality training energy data. To address this, we advocate for greater transparency and accessibility in profiling practices. Ideally, publicly released models should be accompanied by their energy and power profiling logs. Lowering the technical barrier to profiling---e.g., through \gls{dl} frameworks that natively support energy tracking or profilers requiring minimal code changes---could accelerate this process. Additionally, existing tools such as CodeCarbon could be improved by offering finer-grained data. Currently, CodeCarbon provides only cumulative energy values, limiting insight into temporal patterns and model behavior over training time.

To mitigate data scarcity, our findings suggest generating synthetic profiling data as a promising direction. One could measure a representative segment of stable power draw, observable after a few training epochs, and extrapolate a full energy profile using noise-injected replication. This would reduce the need for exhaustive training while still producing valuable input for pre-training energy estimators, saving time, computational resources, and energy.

\section{Threats to Validity}\label{sec:threats}
We identified and addressed several threats to validity in the design and execution of this study. Below, we outline the primary concerns and our mitigation strategies.

A key threat to conclusion validity lies in the potential low statistical power of our analyses. To mitigate this, we followed a rigorous data collection and execution protocol (see Section~\ref{sec:execution-plan}) and ensured a substantial sample size. In total, 1,225 experimental runs were conducted, covering various combinations of factors, treatments, and repetitions, thereby increasing the reliability of our statistical inferences.

External factors such as hardware instability or background processes may introduce noise into energy measurements, threatening the reliability of measures. To reduce these effects, we executed each sub-experiment 30 times and ensured a clean execution environment with only essential processes running. The inherent randomness of deep learning training was addressed by fixing the seeds of all pseudo-random number generators, as recommended by the Keras documentation.\footnote{\href{https://keras.io/getting_started/faq/\#how-can-i-obtain-reproducible-results-using-keras-during-development}{https://keras.io/getting\_started/faq}} We also randomized the execution order of runs to minimize the impact of any temporal anomalies.

A mono-method bias could affect construct validity, as we used single metrics for computational complexity and model accuracy. For complexity, we relied on \gls{flops}, which is widely used in efficient \gls{dl} research~\cite{Neshatpour2018, Boutros2021}. For accuracy, we used the $F_1$ score due to its ability to balance precision and recall in a single, interpretable metric. Other performance metrics, such as accuracy, precision, recall, and AUC, were also recorded and are available in our \textbf{replication package}~\cite{replicationPackage}.

Concerning internal validity, successive experiment runs may introduce a history bias, such as increased hardware temperature influencing results. To mitigate this, we introduced a 5-minute cooldown period between runs and performed an initial warm-up task to ensure consistent hardware conditions across all executions.

Finally, for external validity, our findings are based on a limited set of five \gls{dl} architectures and three training environments. While these choices reflect a range of real-world settings, from entry-level to advanced, we acknowledge that the generalizability of our results may be constrained. Nonetheless, the selected models are widely used and representative, providing a solid foundation for future work.

Another external validity threat concerns the generalizability of the results of the energy estimation methods. Both the STEP and PRE methods were trained and evaluated using data from a limited set of models, datasets, and training environments. While our findings demonstrate strong predictive performance within this scope, their accuracy may vary when applied to different architectures, hardware configurations, or larger-scale models. Although we included multiple regression algorithms and used cross-validation to enhance robustness, generalization to unseen setups is not guaranteed. Further studies using diverse profiling data are needed to confirm the broader applicability of these estimators.

Lastly, the specific datasets and contexts used may also impact generalizability. Results could vary with different datasets or data volumes. To account for this, we report normalized energy consumption metrics (see Section~\ref{sec:data-analysis}). However, we recognize that metrics like \gls{gpu} usage and $F_1$ score are inherently problem-dependent and should be interpreted within the dataset context.

\section{Conclusion and Future Work}\label{sec:conclusion}
Over the past decade, the rapid adoption of Deep Learning (\gls{dl}) in computer vision has raised growing concerns about its environmental impact. In this study, we explored how two major design factors~\cite{delreySoftwareDesignDecisions2024}---model architecture and training environment---affect the energy consumption of image classification models during training. We further analyzed the energetic behavior of \gls{dl} training processes and introduced four new data-driven energy estimation methods. Our experimental design encompassed six widely used \gls{dl} architectures, three training environments, and three benchmark datasets.

Our findings show that selecting energy-efficient architectures can reduce training energy consumption by up to 80\%, with negligible losses in accuracy. Crucially, we also demonstrate that the training environment plays a non-trivial role in determining energy efficiency. Our results highlight a significant interaction effect between model computational complexity and hardware capability, suggesting that mismatches in this pairing can lead to substantial energy waste. Therefore, we recommend that training environment constraints be considered alongside model selection.

We also critically examined the reliability of existing energy estimation practices and found them to be biased and often misleading. In response, we proposed the STEP and PRE methods, which achieved substantially lower error rates. These methods provide practitioners with practical tools to assess energy consumption either during or before training. We also discussed their potential use in energy reporting initiatives, such as Hugging Face's AI Energy Score.

From a practitioner's perspective, our work offers actionable insights into designing energy-aware training pipelines, selecting efficient model--hardware pairs, and evaluating the energy impact of training. Similarly, cloud service providers can use our findings to better allocate \gls{gpu} resources and reduce operational carbon footprints by aligning model types with the most suitable hardware.

From a research perspective, our contributions open several avenues for future work. Our results underscore the importance of understanding the interplay between model architecture and \gls{gpu} resource provisioning for improving the sustainability of \gls{ai} systems. Moreover, our findings highlight the pressing need for more accurate, generalizable energy estimation techniques, especially those grounded in runtime profiling rather than static assumptions like \gls{flops} or \gls{tdp}.

This study focused primarily on specific aspects of the training phase. Future work will extend this research in several directions. First, we plan to replicate our experiments in additional training environments such as AWS and Azure to assess generalizability. Second, we aim to explore energy usage across other phases of the DL lifecycle, including hyperparameter tuning and inference. Third, we intend to extend our evaluation to other computer vision tasks such as object detection. Finally, we aim to develop predictive models for estimating the optimal GPU requirements for a given \gls{dl} model to achieve energy-efficient training.

\section{CRediT author contribution}
\textbf{Santiago del Rey:} Conceptualization, Data curation, Formal analysis, Investigation, Methodology, Software, Validation, Visualization, and Writing -- Original Draft, Writing -- Review \& Editing. \textbf{Lu{\'i}s Cruz:} Conceptualization, Methodology, Supervision, and Writing -- Review \& Editing. \textbf{Xavier Franch:} Conceptualization, Funding Acquisition, Methodology, Project Administration, Resources, Supervision, and Writing -- Review \& Editing. \textbf{Silverio Mart{\'i}nez-Fern{\'a}ndez:} Conceptualization, Funding Acquisition, Methodology, Project Administration, Resources, Supervision, and Writing -- Review \& Editing.

\section{Data availability statement}
We provide all the raw and processed energy measurements with the replication package~\cite{replicationPackage}.

\printglossary[type=\acronymtype]

\bibliographystyle{elsarticle-num-names}
\bibliography{bibliography}

\end{document}